%% file: main.tex
\documentclass{article}
\usepackage{spconf,amsmath,graphicx}
\usepackage{booktabs}
\usepackage[table]{xcolor}
\usepackage{microtype}
\usepackage[hidelinks]{hyperref}
\title{FORECAST WORKFLOW BENCH: EVALUATING LANGUAGE-MODEL DECISIONS\\
WITH BUDGETED FORECAST TOOLS}
\name{Shunya Nagashima}
\address{Neurogica Inc.}
\begin{document}
\maketitle
\begin{abstract}
Time-series foundation models (TSFMs) provide forecasts for operational
decisions, but accuracy alone does not determine their value. Evaluating
agents that use these models requires measuring decision quality and
forecast cost. FWBench evaluates this capability on 1,251 electricity and
cycle-hire cases using fixed forecast tools and simulated capacity contracts.
Agents select models, histories and horizons, then submit capacities to
minimize a stated loss--cost objective. We evaluated two hosted and eight local configurations,
including small language models, and tested local models with and without TSFMs. GPT-6 Astra bought
inexpensive short-horizon forecasts selectively, using 2.5\% of the budget,
and outperformed fixed policies when the saved decisions were scored with three loss--cost weightings. FWBench enables reproducible evaluation of how language models select and use
time-series forecasts to make decisions under cost constraints.
\end{abstract}
\begin{keywords}
Benchmarks, agents, large language models, time-series forecasting, decision-making
\end{keywords}
\begin{figure}[!t]
\centering
\includegraphics[width=\columnwidth]{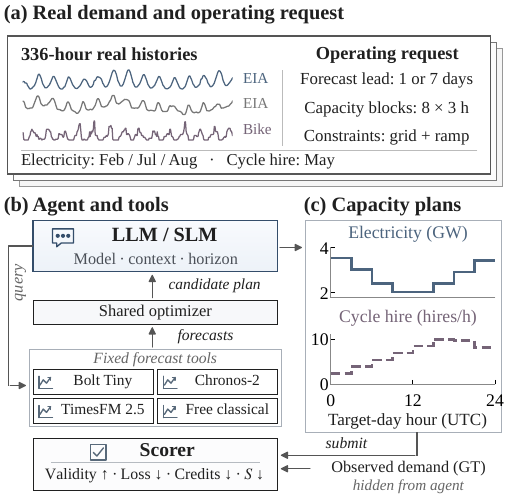}
\caption{FWBench inputs (a), forecasting tools, optimizer and scorer (b),
and Astra's submitted capacities (c): BANC, July 24; Hop Exchange, May 11.
Capacities use physical units; histories show 3-hour medians.}
\label{fig:overview}
\end{figure}
\input{sections/01_introduction}
\input{sections/02_related_work}
\input{sections/03_benchmark}
\input{sections/04_experiments}
\input{sections/05_limitations}
\clearpage
\bibliographystyle{IEEEbib}
\urlstyle{same}
\bibliography{refs}
\end{document}

%% file: sections/01_introduction.tex
\section{Introduction}
Effective decision-making requires anticipating future conditions.
Time-series foundation models (TSFMs) provide accurate forecasts across
domains~\cite{ansari2025chronos2,das2024timesfm}; large language models (LLMs) can
plan with tools~\cite{li2023apibank}.
Existing evaluations examine forecast accuracy and forecaster
selection~\cite{aksu2024gift,ning2026router}, but do not measure whether
a forecast improves capacity decisions enough to justify its cost.

Recent advances have enabled local forecasting with open-weight TSFMs
smaller than one billion parameters~\cite{bolttiny,timesfm25}.
Combined with small language models (SLMs), they could support decisions
without cloud communication. However, existing forecasting benchmarks
have not established how well these local systems make cost-constrained
decisions or how they compare with proprietary LLMs.
Forecast Workflow Bench (FWBench; Fig.~\ref{fig:overview}) evaluates
these capabilities using fixed forecasters, real demand and simulated capacity contracts.

The code\footnote{\urlstyle{same}\url{https://github.com/Neurogica/forecast-workflow-bench}}
and dataset\footnote{\urlstyle{same}\url{https://huggingface.co/datasets/Neurogica/forecast-workflow-bench}}
are publicly available for reproducing the evaluation.
Our contributions are (i) a 1,251-case electricity and cycle-hire benchmark
for evaluating forecast selection, its cost and the resulting capacity decisions;
(ii) comparisons of ten hosted and local configurations with fixed and adaptive
policies; and (iii) paired ablation studies of TSFM access and thinking,
oracle comparisons and sensitivity to objective weights.

%% file: sections/02_related_work.tex
\section{Related Work}
Recent work on time-series foundation models has explored transferring
pretrained knowledge across tasks~\cite{liang2024survey}.
Benchmarks evaluate forecast accuracy~\cite{aksu2024gift}, forecaster
selection~\cite{ning2026router} and broader time-series tasks.
For agent-based systems, TimeSeriesGym evaluates time-series
engineering~\cite{cai2025gym}, while TemporalBench assesses contextual
prediction and event-informed reasoning~\cite{weng2026temporal}.

Recent language-model agents combine planning, memory and external
tools~\cite{wang2024agentsurvey}. Benchmarks test how they use these tools
and whether they complete tasks within resource limits.
Existing benchmarks assess API planning and calling~\cite{li2023apibank},
policy compliance~\cite{yao2024taubench}, stateful tool use~\cite{lu2025toolsandbox}
and industrial operations~\cite{patel2025assetops}.
CostBench measures planning cost gaps~\cite{liu2025costbench}, while
EcoAgent-Bench evaluates budgeted economic decisions~\cite{wu2026ecoagent}.

FWBench evaluates how forecast selection affects capacity decisions and
cost. Hosted and local models use the same fixed tools, capacity constraints
and forecast budget.

%% file: sections/03_benchmark.tex
\section{Benchmark Design}
\subsection{Data construction and observation windows}
FWBench uses U.S. Energy Information Administration form EIA-930 electricity
demand~\cite{eia930,eiareuse} and Transport for London (TfL) cycle-hire
departures~\cite{tflcycles}. Windows must have unique timestamps and pass provider quality checks.
We exclude forecast, adjusted, imputed and nonpositive electricity values,
but retain zero cycle-hire counts.

Agents observe hourly histories and submit capacities one or seven days
before the target day. Timestamps mark the start of each UTC hour; agents
cannot access future demand (ground truth, GT). Cycle-hire capacities
represent departures per hour; the task does not model inventory or rebalancing.

\subsection{Actions and decision loss}
Forecast error alone does not measure the quality of capacity decisions.
We score submitted capacities against observed demand using a decision loss
that penalizes unused capacity and unmet demand under the simulated contract.

For $T$ target hours $h=0,\ldots,T-1$, $y_h$ is observed demand and
$z_h=y_h/s$ demand normalized by the visible-history mean $s>0$.
Plan $a$ contains capacities $a_b$ for $T/d$ consecutive $d$-hour blocks
($b=0,\ldots,T/d-1$; $d$ divides $T$). Feasible actions satisfy
$a_b\in\mathcal G$ and $|a_b-a_{b-1}|\leq r$ for $b\geq1$,
where $\mathcal G$ contains allowed normalized capacities and $r$ limits
changes between adjacent blocks.
Submissions use physical capacities $s a_b$. For demand vector $z=(z_h)$, loss is
\begin{equation}
\ell(a,z)=\frac1T\sum_{h=0}^{T-1}
\left[\alpha(\bar a_h-z_h)_+
+w_h(z_h-\bar a_h)_+\right],
\label{eq:loss}
\end{equation}
where $\bar a_h=a_{\lfloor h/d\rfloor}$ is normalized capacity at hour $h$,
$(\cdot)_+$ clips negative values to zero, and $\alpha$ and $w_h$ are the
surplus and hourly shortage weights supplied to the agent.
Submissions require unique block IDs, finite grid values and feasible ramps;
invalid answers are penalized above every feasible loss.

\subsection{Forecast tools and decision support}
Table~\ref{tab:tools} lists the tools, including fixed Chronos-Bolt Tiny,
Chronos-2 and TimesFM 2.5 services~\cite{ansari2025chronos2,das2024timesfm,timesfm25,bolttiny}.
Unaffordable requests are refused. A shared optimizer uses dynamic programming to minimize expected contract
loss given classical or TSFM forecasts.
Agents can test past forecast accuracy through replay: a model predicts
from a historical origin using only earlier observations. The tool returns
pinball loss on observed hours and charges the forecast tariff.
\input{tabs/available_tools}

Prompts provide timestamps, block IDs, units, contract terms and the
objective below. Agents submit capacities by block ID; browsing and file
access are disabled.

\subsection{Cost accounting, objective and references}
CPU inference times determine fixed charges by model, history length and
forecast horizon. Credits exclude language-model inference, loading and
optimization; they do not measure evaluation runtime.

To balance decision quality and forecast cost, case $i$ in domain $j$
with credit budget $B$ receives score
\begin{equation}
 S_i=\tfrac12\left[\frac{\ell_i-F_i}{\sigma_j}+\frac{c_i}{B}\right],
\label{eq:score}
\end{equation}
where $\ell_i$ is loss from Eq.~\eqref{eq:loss} or the invalid penalty,
$c_i$ is the credit charge, and $F_i$ is the loss obtained by optimizing
capacities with the actual target demand known.
The scale $\sigma_j>0$ is the hour-of-day policy's mean excess loss on
pre-evaluation replay cases. This policy uses past observations at each
target hour. Prompts omit $F_i$ because it requires unseen demand;
this constant does not affect the optimal action.
Equal loss--cost weights express a preference, not an economic exchange rate
(Section~\ref{sec:robustness}). Normalized excess loss is
$R_i=(\ell_i-F_i)/\sigma_j$; lower $R_i$ and $S_i$ are better.
We average $R_i$, $S_i$, loss and credits by series and lead, then average
these means equally within each cohort. Cohort weights
(Section~\ref{sec:track}) give the reported $R$, $S$, loss and credits $c$.
Invalid submissions retain their loss penalties in these averages.

\input{tabs/objective_main_results}

Baselines without language models include fixed policies that use one
method for all cases, and selectors that choose free methods by 24-hour
backtest loss.
Selectors break ties by name and refit on full histories.

Two oracles measure the score achievable by selecting plans with true
future demand: the catalog oracle minimizes $S_i$ over affordable
full-history plans from eight classical and three TSFM methods, paying only
for its choice; the free-menu oracle excludes TSFMs.

\begin{figure}[t]
\centering
\includegraphics[width=\columnwidth]{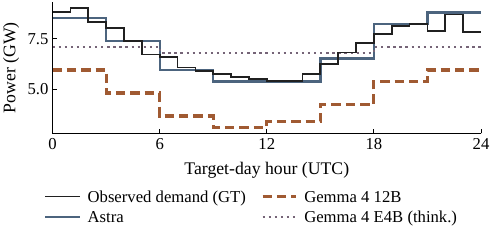}
\caption{SRP, July 25, one-day lead: demand and capacities.
Case fixed before the run; configurations selected post hoc.}
\label{fig:qualitative}
\end{figure}

%% file: tabs/available_tools.tex
\begin{table}[t]
\centering
\caption{Tools available to all agents. Forecast charges follow the selected model, history and horizon; other tools are free.}
\label{tab:tools}
\smallskip
\small
\begin{tabular*}{\columnwidth}{@{}p{.27\columnwidth}@{\extracolsep{\fill}}p{.69\columnwidth}@{}}
\toprule
Operation & Information returned \\
\midrule
Inspect & Models, history coverage, remaining budget \\
Read history & Observed hourly demand before the cutoff \\
Quote & Cost of the requested forecast \\
Forecast and plan & Forecast metadata and optimized capacities \\
Classical plan & Capacities from a free historical method \\
Replay forecast & Past-only forecast error and scored coverage \\
\bottomrule
\end{tabular*}
\end{table}

%% file: tabs/objective_main_results.tex
\begin{table}[t]
\centering
\caption{1,251 cases, full tools. Valid: feasible plans; $c$: credits. $S$ includes invalid penalties; $S_v$ excludes them and renormalizes weights over differing valid subsets. $\dagger$: $S_v$ from only three cases. T: thinking; bold: agents.}
\label{tab:agents}
\smallskip
\small
\setlength{\tabcolsep}{1.5pt}
\begin{tabular*}{\columnwidth}{@{\extracolsep{\fill}}lrrrrr@{}}
\toprule
Configuration & Valid$\uparrow$ & Loss$\downarrow$ & $c$$\downarrow$ & $S\downarrow$ & $S_v\downarrow$ \\
\midrule
\rowcolor{black!8}
\multicolumn{6}{l}{\textit{SLMs}} \\
Gemma 4 E4B~\cite{gemma2026} & 1019 & 0.7074 & 74 & 4.6294 & 1.187 \\
Qwen3 4B~\cite{yang2025qwen3} & 1242 & 0.7175 & \textbf{0} & 4.5478 & 4.113 \\
Qwen3 4B\textsuperscript{T} & 1237 & 0.5196 & \textbf{0} & 2.5232 & 2.429 \\
Gemma 4 E4B\textsuperscript{T} & \textbf{1251} & 0.3649 & 1209 & 1.0465 & 1.046 \\
\rowcolor{black!8}
\multicolumn{6}{l}{\textit{Larger local models}} \\
Qwen3.6 35B-A3B~\cite{qwen36moe} & 3$^{\dagger}$ & 2.0080 & 1176 & 17.9773 & 0.544 \\
Qwen3.6 35B-A3B\textsuperscript{T} & 160 & 1.8533 & 1464 & 16.4632 & 1.003 \\
Gemma 4 12B\textsuperscript{T}~\cite{gemma2026} & 490 & 1.2837 & 58104 & 10.6358 & 1.151 \\
Gemma 4 12B & 766 & 0.9611 & 46091 & 7.3281 & 0.983 \\
\rowcolor{black!8}
\multicolumn{6}{l}{\textit{Hosted LLMs}} \\
GPT-5.4 nano~\cite{openai54small} & 1139 & 0.5160 & 2886 & 2.6411 & 0.893 \\
GPT-6 Astra~\cite{openaiastra} & \textbf{1251} & \textbf{0.3044} & 2802 & \textbf{0.3964} & \textbf{0.396} \\
\midrule\midrule
\rowcolor{black!8}
\multicolumn{6}{l}{\textit{Policies without language models}} \\
Fixed Chronos-2~\cite{ansari2025chronos2} & 1251 & 0.3050 & 108783 & 0.8651 & -- \\
Backtest (8 windows) & 1251 & 0.3265 & 0 & 0.6175 & -- \\
Backtest selector & 1251 & 0.3259 & 0 & 0.6109 & -- \\
Fixed Chronos-Bolt Tiny~\cite{bolttiny} & 1251 & 0.3087 & 39100 & 0.6014 & -- \\
Weekday empirical & 1251 & 0.3220 & 0 & 0.5726 & -- \\
Lead-matched selector & 1251 & 0.3211 & 0 & 0.5633 & -- \\
Hour-of-day empirical & 1251 & 0.3165 & 0 & 0.5147 & -- \\
\midrule
\rowcolor{black!8}
\multicolumn{6}{l}{\textit{Hindsight references (not agents)}} \\
Catalog oracle & -- & 0.2899 & 1053 & 0.2393 & -- \\
Perfect demand & -- & 0.2676 & -- & 0.0000 & -- \\
\bottomrule
\end{tabular*}
\end{table}

%% file: sections/04_experiments.tex
\section{Experiments}
\subsection{Experimental settings}
\label{sec:track}
We set $T=24$, $d=3$ (eight actions), $\alpha=0.3$,
$\mathcal G=\{0.40,0.45,\ldots,2.00\}$, and $r=0.30$.
Shortage weights followed six-hour segments $1,1,2,1.5$, shifted by target
weekday. Each case allowed $B=113{,}726$ forecast credits and 12 tool calls. Prompts instructed agents to minimize $S$ in Eq.~\eqref{eq:score}.
We set $\sigma_j=0.0542$ for electricity and $\sigma_j=0.0434$ for cycle hire. Selectors used four or eight short-lead replays, or lead-matched replays
with shorter histories.

We used 1,091 electricity cases from 50 authorities in February, July and
August 2026, and 160 cycle-hire cases from 20 stations in May 2026.
Cohorts comprised three electricity seasons (weight 1/6 each) and
transport (1/2).
The target day began one or seven days after the decision. Agents had
336 hourly observations ending three hours before the decision and chose
how much history to use. Forecasts began at the next hourly timestamp;
50 or 194 steps covered the target day.

The two hosted models used high reasoning effort; the four quantized local
models were tested with thinking off and on (Table~\ref{tab:agents}).
SLMs had $<10$B total parameters; larger local models had $\geq10$B.
In total, 22,518 conversations were scored.
Post-hoc normalization accepted equivalent JSON and quoted numbers
without changing numerical values.

\input{tabs/local_resources}
Table~\ref{tab:resources} profiles one case per cohort--lead pair on an RTX
PRO 6000 Blackwell with CPU tools. All submissions were valid and used no forecast credits.
GPU energy included idle periods; host memory used proportional set size
(PSS). Bolt on one CPU thread took 7.5/65.9\,ms for 50/194-step forecasts
with 0.45\,GiB peak PSS; CPU energy was not measured.

\subsection{Quantitative results}
\begin{figure}[t]
\centering
\includegraphics[width=\columnwidth]{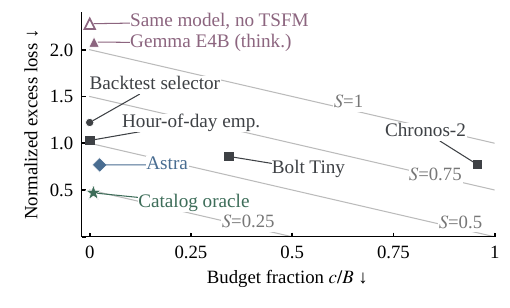}
\caption{Loss--cost trade-off on 1,251 cases; lower left is better.
Diagonals: equal $S$; filled/hollow triangles: Gemma 4 E4B with thinking,
with/without TSFMs; star: catalog oracle.}
\label{fig:frontier}
\end{figure}

\noindent\textbf{Overall comparison.} Table~\ref{tab:agents} shows decision loss, forecast credits and $S$ for
agents and reference policies. Astra achieved similar loss to fixed
Chronos-2 using 2.6\% of its forecast credits. It bought only short-horizon
Bolt Tiny forecasts (Fig.~\ref{fig:frontier}). It ranked first in $S$ overall
and in all eight cohort--lead subsets by point estimate.
Astra's improvements in $S$ and loss over the best free policy and three
selectors had paired 95\% cluster-bootstrap intervals excluding zero.

\noindent\textbf{Local decision quality and completion.} All local configurations
scored worse than hour-of-day empirical. Gemma 4 E4B with thinking
completed every case yet scored more than twice as high, identifying
forecast selection and use as improvement targets beyond valid output. Valid-only $S_v$ excludes failures but
compares different subsets. Qwen3.6 without thinking exhausted the 12-call
limit in 1,245 cases, so workflow non-completion dominated its score.

\subsection{TSFM access and thinking}
\label{sec:formats}
\input{tabs/tsfm_access_effects}
Table~\ref{tab:access} shows no significant score benefit from TSFM access
for any local configuration. Access significantly increased $S$ for Gemma 4
12B without thinking and Qwen3.6 with thinking (paired 95\% intervals
excluded zero). On cases valid with and without TSFMs, access raised $S$
for Gemma (714 cases) but lowered it for Qwen (58 cases).
Thus, Gemma's worse score was not explained by failed submissions alone.
Qwen's lower score on this subset did not establish a benefit across all
cases, which included failures.

With all tools available, thinking lowered $S$ for three models but raised
it for Gemma 4 12B; all four paired intervals excluded zero.
This comparison measured how agents used TSFMs, rather than TSFM
forecast accuracy.

\subsection{Effect of the stated objective}
We tested whether including forecast cost in the objective changed purchases.
Astra and nano received either loss-only instructions or instructions to
minimize the combined loss--cost score $S$ (Eq.~\eqref{eq:score}) on the
same 16 cases, with identical tools and limits. Relative to loss-only instructions, minimizing $S$
reduced Astra's credits by 98.5\% with 1.6\% higher decision loss.
All Astra plans were valid. Only Astra showed a significant reduction in $S$.
The change in purchases under identical tools and budgets indicates that
the stated objective influenced how Astra used forecasts.

In Fig.~\ref{fig:qualitative}, Astra bought a Bolt Tiny forecast after three
free plans. Gemma 4 12B without thinking spent 91\% of its budget on forecasts
ending before the target, but submitted a free plan.

\subsection{Weight sensitivity and repeated executions}
\label{sec:robustness}
We rescored the saved decisions to test sensitivity to loss--cost weights.
Astra remained first under both alternatives (0.25/0.75 and 0.75/0.25),
with significant improvements over each setting's best fixed policy.
Giving more weight to loss made Bolt Tiny score better than hour-of-day empirical.

We repeated Astra on 100 cases to assess run-to-run stability. Both runs
returned valid plans and ranked first against existing results on those cases.
However, the paired 95\% interval against the best fixed policy included
zero, so the repeated ranking did not establish statistical superiority.

%% file: tabs/local_resources.tex
\begin{table}[t]
\centering
\caption{Resident SLMs, thinking on, on eight common cases. Peak GPU/host PSS; mean case time, GPU power and energy.}
\label{tab:resources}
\smallskip
\small
\setlength{\tabcolsep}{2.5pt}
\begin{tabular*}{\columnwidth}{@{\extracolsep{\fill}}lrrrr@{}}
\toprule
SLM & Memory (GiB) & Time (s) & GPU (W) & GPU (kJ) \\
\midrule
Gemma 4 E4B & 4.60 / 3.69 & 15.9 & 245 & 3.90 \\
Qwen3 4B & 9.84 / 1.06 & 13.9 & 310 & 4.30 \\
\bottomrule
\end{tabular*}
\end{table}

%% file: tabs/tsfm_access_effects.tex
\begin{table}[t]
\centering
\caption{TSFM removal, 1,251 cases. Valid/$S$: no-TSFM; $\Delta S$: full minus no-TSFM (negative favors access). 95\% CI: paired cluster-bootstrap confidence interval; bold excludes zero.}
\label{tab:access}
\smallskip
\small
\setlength{\tabcolsep}{2.5pt}
\begin{tabular*}{\columnwidth}{@{\extracolsep{\fill}}lrrrr@{}}
\toprule
Configuration & Valid$\uparrow$ & $S\downarrow$ & $\Delta S$ & 95\% CI \\
\midrule
\rowcolor{black!8}
\multicolumn{5}{l}{\textit{SLMs}} \\
Gemma 4 E4B~\cite{gemma2026} & 1008 & 4.052 & $+0.577$ & $[-0.277, 1.321]$ \\
Qwen3 4B~\cite{yang2025qwen3} & 1242 & 4.514 & $+0.034$ & $[-0.042, 0.159]$ \\
Qwen3 4B\textsuperscript{T} & 1243 & 2.464 & $+0.059$ & $[-0.117, 0.230]$ \\
Gemma 4 E4B\textsuperscript{T} & 1250 & 1.139 & $-0.092$ & $[-0.197, 0.008]$ \\
\rowcolor{black!8}
\multicolumn{5}{l}{\textit{Larger local models}} \\
Qwen3.6 35B-A3B~\cite{qwen36moe} & 1 & 18.090 & $-0.113$ & $[-0.346, 0.119]$ \\
Qwen3.6 35B-A3B\textsuperscript{T} & 448 & 12.436 & $+4.027$ & {\boldmath$[3.338, 4.732]$} \\
Gemma 4 12B\textsuperscript{T}~\cite{gemma2026} & 584 & 9.662 & $+0.974$ & $[-0.345, 2.184]$ \\
Gemma 4 12B & 1146 & 2.062 & $+5.266$ & {\boldmath$[4.573, 5.944]$} \\
\bottomrule
\end{tabular*}
\end{table}

%% file: sections/05_limitations.tex
\section{Limitations and Future Work}
The main limitation is coverage: we evaluated two domains over 16 dates
under simulated contracts.
Cycle-hire counts excluded unmet demand. Weather could correlate results
across sites. Confidence intervals were conditional on the sampled dates
and were not adjusted for multiple testing.

Fixed tariffs and one calibration week limited economic interpretation.
Hosted comparisons lacked TSFM-removal tests and full-set repeated runs.
The common harness did not evaluate differences between complete agent systems.

In future work, we plan to evaluate task-specific decision models, extend
temporal and domain coverage, compare complete agent systems with their
own harnesses, and repeat TSFM-access comparisons across calibration periods. We will measure end-to-end latency, memory
and energy on edge devices while retaining common tools and scoring.
Updated tool catalogs will require new oracle references.

\section{Conclusion}
FWBench evaluates LLM and SLM decisions with fixed forecast tools and a
common loss--cost objective. Astra's selective purchasing achieved similar
loss to fixed Chronos-2 at lower forecast cost, while fully valid SLM
plans still lagged behind free policies. Better forecast selection could narrow the gap to the hindsight catalog oracle;
stronger TSFMs could further improve decision quality. FWBench supports
evaluating improvements in forecast selection and decision quality for
both LLMs and SLMs under common tools and cost constraints.